\documentclass{article}

\usepackage{microtype}
\usepackage{graphicx}
\usepackage{subfigure}
\usepackage{booktabs} 

\usepackage{multirow}
\usepackage{longtable}
\usepackage{amsmath}
\usepackage{tabularx}
\usepackage{url}

\usepackage{amsmath,bm}
\usepackage{amsfonts}

\def\eqref#1{equation~\ref{#1}}

\def\1{\bm{1}}

\def\vd{{\bm{d}}}

\def\vh{{\bm{h}}}

\def\vp{{\bm{p}}}

\def\vs{{\bm{s}}}

\def\vx{{\bm{x}}}
\def\vy{{\bm{y}}}

\def\evw{{w}}

\def\mH{{\bm{H}}}

\DeclareMathAlphabet{\mathsfit}{\encodingdefault}{\sfdefault}{m}{sl}
\SetMathAlphabet{\mathsfit}{bold}{\encodingdefault}{\sfdefault}{bx}{n}
\usepackage{hyperref}

\usepackage[accepted]{icml2021}

\usepackage{icml2021}

\usepackage[noabbrev,capitalize]{cleveref} 

\crefname{equation}{equation}{equations}   
\crefname{footnote}{footnote}{footnotes}   
\crefname{line}{line}{lines}               
\crefname{section}{\S}{\S\S}
\Crefname{section}{\S}{\S\S}    
\crefformat{section}{#2\S#1#3}  
\Crefformat{section}{#2\S#1#3}
\crefrangeformat{section}{\S\S#3#1#4--#5#2#6}
\Crefrangeformat{section}{\S\S#3#1#4--#5#2#6}

\icmltitlerunning{
Text Generation with Diffusion Language Models: A Pre-training Approach with Continuous Paragraph Denoise}

\newcommand{\ours}{\texttt{GENIE}}
\newcommand{\weizhu}[1]{\textcolor{red}{#1 }}

\begin{document}

\twocolumn[
\icmltitle{
Text Generation with Diffusion Language Models: A Pre-training Approach with Continuous Paragraph Denoise
}



\icmlsetsymbol{equal}{*}

\begin{icmlauthorlist}
\icmlauthor{Zhenghao Lin}{xmu,Intern}
\icmlauthor{Yeyun Gong}{msra}
\icmlauthor{Yelong Shen}{microsoft}
\icmlauthor{Tong Wu}{tsinghua,Intern}
\icmlauthor{Zhihao Fan}{fudan,Intern}\\
\icmlauthor{Chen Lin}{xmu}
\icmlauthor{Nan Duan}{msra}
\icmlauthor{Weizhu Chen}{microsoft}
\end{icmlauthorlist}

\icmlaffiliation{xmu}{Xiamen University}
\icmlaffiliation{tsinghua}{Tsinghua University}
\icmlaffiliation{fudan}{Fudan University}
\icmlaffiliation{msra}{Microsoft Research Asia} \icmlaffiliation{microsoft}{Microsoft}  
\icmlaffiliation{Intern}{This work was done during an internship in MSRA}

\icmlcorrespondingauthor{Chen Lin}{chenlin@xmu.edu.cn}

\icmlkeywords{Natural Language Generation, Pretraining}

\vskip 0.3in
]

\printAffiliationsAndNotice{}




\begin{abstract}

In this paper, we introduce a novel d\textbf{I}ffusion language mod\textbf{E}l pre-training framework for text generation, which we call \textbf{\ours{}}. \ours{} is a large-scale pretrained diffusion language model that consists of an encoder and a diffusion-based decoder, which can generate text by gradually transforming a random noise sequence into a coherent text sequence. To pre-train \ours{} on a large-scale language corpus, we design a new \textit{continuous paragraph denoise} objective, which encourages the diffusion-decoder to reconstruct a clean text paragraph from a corrupted version, while preserving the semantic and syntactic coherence. We evaluate \ours{} on four downstream text generation benchmarks, namely~\textsc{XSum},~\textsc{CNN/DailyMail},~\textsc{Gigaword}, and~\textsc{CommonGen}. Our experimental results show that \ours{} achieves comparable performance with the state-of-the-art autoregressive models on these benchmarks, and generates more diverse text samples. The code and models of \ours{} are available at \url{https://github.com/microsoft/ProphetNet/tree/master/GENIE}.





\end{abstract}

\section{Introduction}\label{sec.introduction}

Text generation is a crucial task in natural language processing, which aims to produce fluent and coherent texts for various applications. Previous text generation methods mainly relied on recurrent neural networks (RNNs)~\cite{pawade2018story,song2018graph,gu2016incorporating,qi2021bang}, which generate texts sequentially from left to right. However, RNNs suffer from issues such as long-term dependency and exposure bias. Recently, Transformer~\cite{vaswani2017attention}, a self-attention based neural network, has emerged as the dominant paradigm for text generation, thanks to its ability to capture global dependencies and leverage large-scale pre-trained language models~\cite{qi2020prophetnet,lewis2019bart,2020t5}. Transformer-based methods typically adopt an encoder-decoder architecture, where the encoder maps the input text to a sequence of hidden vectors, and the decoder generates the output text either autoregressively (AR) or non-autoregressively (NAR). AR decoding is more accurate but slower, as it predicts each word conditioned on the previous ones. NAR decoding is faster but less precise, as it predicts all words simultaneously without modeling the dependencies among them.


In this paper, we present a new text generation approach, called \ours{}, that integrates the diffusion model and Transformer-based method. The diffusion model is a generative model that reverses a stochastic process of adding noise to the data, and has shown promising results in image~\cite{ho2020denoising,song2020denoising}, molecule~\cite{hoogeboom2022equivariant}, video~\cite{ho2022video}, and text~\cite{li2022diffusion,gong2022diffuseq,strudel2022self,reid2022diffuser} generation. \ours{} follows the encoder-decoder architecture, where the encoder transforms the input text to hidden vectors, and the diffusion model restores the output text from a random Gaussian noise, guided by the encoder hidden vectors. The diffusion model iterates over multiple time steps, and gradually denoises the output text at each step.


To leverage the large-scale unlabeled text data, we also propose an end-to-end pre-training method for \ours{}. Unlike the existing pre-training tasks that involve masking or splitting tokens or texts~\cite{qi2020prophetnet,lewis2019bart,2020t5}, we design a novel pre-training task, called \textit{continuous paragraph denoise} (CPD). CPD requires the model to predict the noise added to continuous paragraphs in the current time step, given the paragraph context and the noisy paragraph information.


We evaluate \ours{} on four popular text generation benchmarks: XSum~\cite{NarayanCL18}, CNN/DailyMail~\cite{HermannKGEKSB15}, Gigaword~\cite{RushCW15}, and CommonGen~\cite{lin2019commongen}. The experimental results demonstrate that \ours{} achieves competitive performance with Transformer-based AR methods, and that the proposed pre-training method can effectively improve the performance. We notice that ~\ours{} has achieved significant improvements in diversity metric. To evaluate the multiple outputs of the generation model, we design an automatic annotation method based on large language model. We also conduct ablation studies to analyze the impact of the diffusion steps and pre-training steps.


The main contributions of this work are summarized as follows:
\begin{itemize}
    \item We propose \ours{}, the first large-scale language pre-trained model based on the diffusion framework, which can generate high-quality texts for sequence-to-sequence tasks.
    \item We introduce a novel CPD loss as the pre-training objective, which can enhance the model's ability to denoise noisy texts and capture the paragraph-level coherence.
    \item We validate the effectiveness of the pre-trained diffusion model on downstream tasks, and design a new automatic annotation method for the evaluation based on large language model. We also provide extensive analyses on the model's behavior and properties.
\end{itemize}

\section{Preliminary}
\subsection{Task Definition}
In the classical sequence-to-sequence task, given a source text $\vs = \{\evw^s_1, \evw^s_2, \dots, \evw^s_n \}$ with $n$ tokens, it generates target text sequence $\vy = \{\evw^y_1, \evw^y_2, \dots, \evw^y_n \}$. A sequence generation model can achieve this by modeling the conditional probability: $p\left(\vy \mid \vs \right)$.
\comment{\weizhu{Given a source text $\vs = \{\evw^s_1, \evw^s_2, \dots, \evw^s_n \}$ with $n$ tokens, our goal is to produce a target text $\vy = \{\evw^y_1, \evw^y_2, \dots, \evw^y_n \}$. A non auto-regressive~(NAR) generation model can achieve this by:}}
\subsection{Diffusion model}
In the diffusion model, the diffusion process can be regarded as a discrete-time Markov process. The diffusion process starts with initial state $\vx_0$ at time step $t=0$, where $\vx_0$ is the Gaussian distribution of the original data. It gradually adds Gaussian noises to $\vx_0$ in the forward diffusion process according to a variance schedule $\beta_1, ..., \beta_T$. At the time step $t+1$, the latent variable $\vx_{t+1}$ is only determined by the $\vx_t$ at time $t$, expressed as:
\begin{equation}
\footnotesize
q\left(\vx_{t+1} \mid \vx_{t}\right)=\mathcal{N}\left(\vx_{t+1} ; \sqrt{1-\beta_{t+1}} \vx_{t}, \beta_{t+1} \mathbf{I}\right)
\label{equ:forwarddiff}
\end{equation}
As $t$ increases, $\vx_t$ becomes closer to standard Gaussian noise $\mathcal{N}(\vx_{T};0,\mathbf{I})$.

\comment{ \weizhu{
The diffusion model simulates a Markov process of gradually adding Gaussian noise to a variable $\vx_0$ over time steps $t=0,1,\ldots,T$. At each time step $t+1$, the latent variable $\vx_{t+1}$ is sampled from a Gaussian distribution with mean $\sqrt{1-\beta_{t+1}} \vx_{t}$ and covariance $\beta_{t+1} \mathbf{I}$, where $\beta_{t+1} \in (0,1)$ controls the noise level:
\begin{equation}
\footnotesize
q\left(\vx_{t+1} \mid \vx_{t}\right)=\mathcal{N}\left(\vx_{t+1} ; \sqrt{1-\beta_{t+1}} \vx_{t}, \beta_{t+1} \mathbf{I}\right)
\label{equ:forwarddiff}
\end{equation}
As $t$ increases, $\vx_t$ becomes closer to standard Gaussian noise $\mathcal{N}(\vx_{T};0,\mathbf{I})$.
}

xusong    }

The diffusion model learns to perform the inverse diffusion process during generation, which predicts the noise given the current state $\vx_{t}$ at time step $t$. The previous state $\vx_{t-1}$ can be reconstructed by subtracting the noise and rescaling the mean. Thus, the distribution of $\vx_{t-1}$ given $\vx_t$ is a Gaussian with mean $\mu^{t-1}_{\theta}$ and variance ${{\sigma}^{t-1}_{\theta}}^2$:
\begin{equation}
\footnotesize
p\left(\vx_{t-1} \mid \vx_{t} \right) =  \mathcal{N}\left(\vx_{t-1} ;  \mu^{t-1}_{\theta} , {\sigma}^{t-1}_{\theta}\right)\label{equ:inversedprob}
\end{equation}
\begin{equation}
\footnotesize
\mu^{t-1}_{\theta}=\frac{1}{\sqrt{\alpha_{t}}}\left(\vx_{t}-\frac{\beta_{t}}{\sqrt{1-\bar{\alpha}_{t}}} z_{\theta}\left(\vx_{t}, t\right)\right)
\label{equ:inversediffmean}
\end{equation}
\begin{equation}
\footnotesize
{{\sigma}^{t-1}_{\theta}}^2=\frac{1-\bar{\alpha}_{t-1}}{1-\bar{\alpha}_{t}} \cdot \beta_{t} \quad
\label{equ:inversediffvar}
\end{equation}
where $\alpha_t = 1 - \beta_t$, $\bar{\alpha}_{t} = \prod_{i=1}^t \alpha_i$ and $z_{\theta}$ is predicted by a neural network parameterized by $\theta$. The diffusion model is trained by minimizing the mean squared error between $\mu^{t-1}_{\theta}$ and the true mean $\hat{\mu}_{t-1}$, which is computed from the reverse conditional distribution $q(\vx_{t-1}|\vx_{t},\vx_{0})$:
\begin{equation}
\footnotesize
q\left(\vx_{t-1} \mid \vx_{t}, \vx_{0}\right)=\mathcal{N}\left(\vx_{t-1} ; \hat{\mu}_{t-1}, \hat{\beta}_{t-1} \mathbf{I}\right)
\label{equ:reconddiff}
\end{equation}
\begin{equation}
\footnotesize
\hat{\mu}^{t-1}_{\theta}=\frac{\sqrt{\bar{\alpha}_{t-1} }\beta_{t}}{ 1 - \bar{\alpha}_t } \vx_0 + \frac{\sqrt{\alpha_t} \left( 1 - \bar{\alpha}_{t-1} \right) }{1-\bar{\alpha}_t} \vx_t 
\label{equ:reconddiff_mu}
\end{equation}
Following the variational lower bound (VLB) approach~\cite{ho2020denoising}, the diffusion model can be trained by minimizing the loss function:
\begin{equation}
\footnotesize
\mathcal{L}_{\text {\emph{diff}}}=\sum_{t=1}^{T} \underset{q\left(\vx_{t} \mid \vx_{0}\right)}{\mathbb{E}}\left\|\mu^{t-1}_{\theta}-\hat{\mu}_{t-1}\right\|^{2}
\label{equ:diffobject}
\end{equation}

\section{Model}\label{sec.model.structure}

\begin{figure*}[t]
\centering
\includegraphics[width=0.8\textwidth]{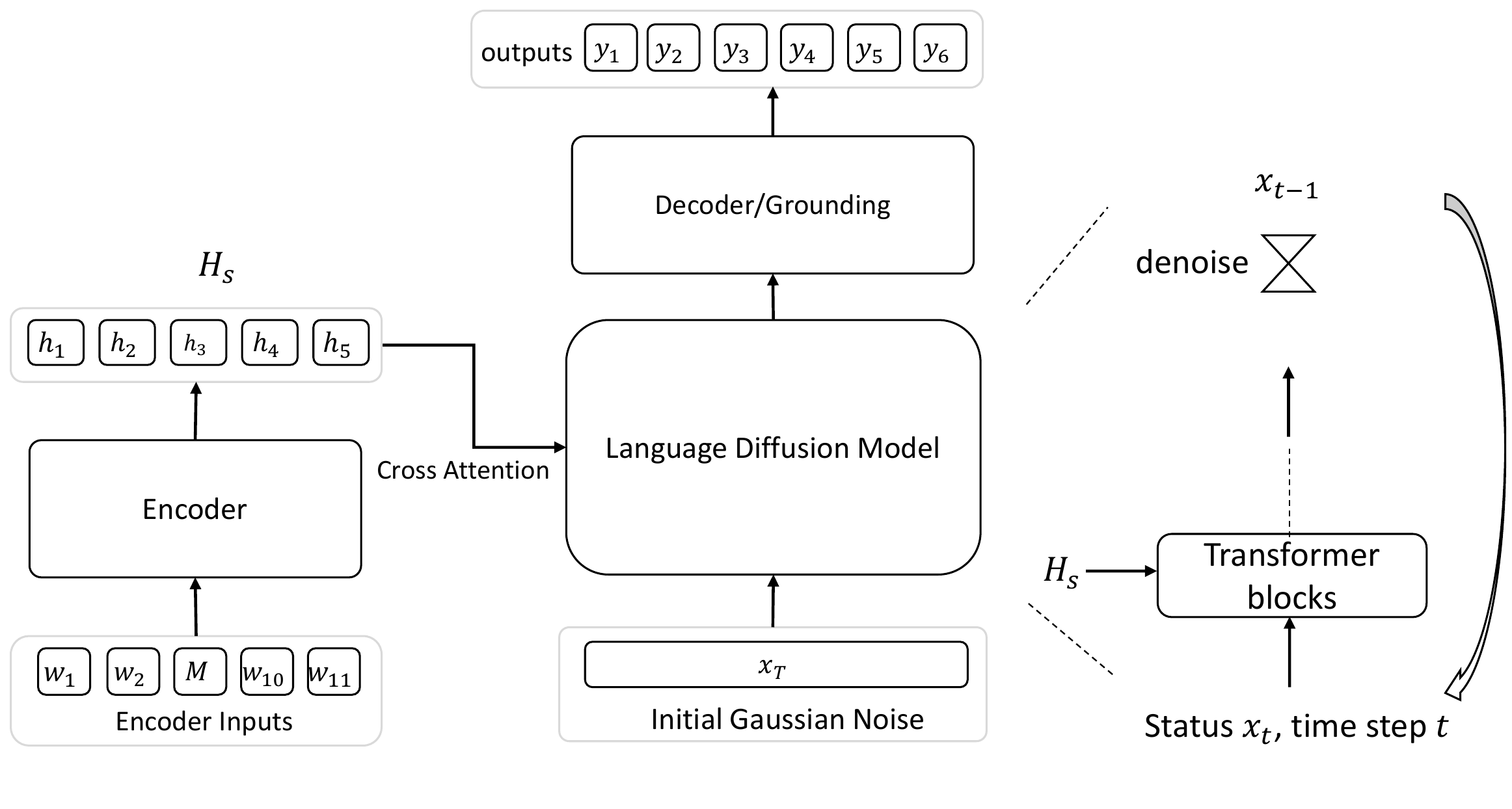}
\caption{The framework of \ours{}. 
We take the masked source sequence $\vs$ as the input of the Encoder to obtain the hidden information $\mH_{\vs}$,  and interact with Language Diffusion Model through cross attention. Language Diffusion Model restores the randomly initial Gaussian noise to the output text $\vy$ through the iterative denoising and grounding process.
}
\label{fig:pipeline}
\end{figure*}




\ours{} is the proposed diffusion language model for pretraining, it adopts the sequence-to-sequence framework as illustrated in \cref{fig:pipeline}. \ours{} could generate a high-quality text sequence $\vy$  given a source text $\vs$, such as producing $\vy:$ \emph{Messi's performance} from $\vs:$ \emph{In the World Cup 2022, [MASK] won people's praise.}. To achieve this, \ours{} leverages two components: a bidirectional encoder model and a cross-attention diffusion model. The encoder model encodes the source text $\vs$ into a set of hidden vectors $\mH_{\vs} = \text{Encoder}(\vs)$, which indicates the distributed representation of $\vs$. The diffusion model takes $\mH_{\vs}$ and a Gaussian noise as inputs, and iteratively refines the data by applying a sequence of denoising operations. In contrast to the traditional autoregressive text generation paradigm, which generates one token at a time, the diffusion model in \ours{} outputs the sequence of embeddings in parallel at each denoising step, making \ours{} a non-autoregressive generation~(NAR) model.






\paragraph{Encoder} The encoder in \ours{} is a 6-layer transformer model which takes the source text $\vs$ as input with bidirectional self-attention. 
Specifically, given a source text sequence $\vs = \{\evw^s_1, \evw^s_2, \dots, \evw^s
_n \}$ with $n$ tokens, the encoder model computes the vector $\vh_i$ for each token $\evw_i$. Thus, the source text $\vs$ can be represented as $\mH_{\vs}$ by the encoder model:
\begin{equation}
\footnotesize
\mH_{\vs} = \{\vh_1, \vh_2, ..., \vh_n\} = \text{Encoder}(\vs)
\label{equ:encoderencode}
\end{equation}


\paragraph{Language Diffusion Model} The diffusion model in \ours{} is a 6-layer transformer with cross-attention on the source text representation $\mH_{\vs}$. It learns to predict Gaussian noise $z_{\theta}\left(\vx_{t}, t, \mH_{\vs}\right)$ conditioned on the current diffusion step $t$ and the state $\vx_{t}$, where $\vx_{t}$ is the continuous latent representation of the target text. We use an embedding function and a clamping trick to ground the continuous state $\vx_{t}$ with discrete target tokens, which will be elaborated in the following section.


\paragraph{Inference Phase} To generate text from the diffusion model, we start from the final step $t=T$ and sample a state $\vx_{T}$ from a standard Gaussian distribution. Then we iteratively generate the noise for the previous step using equations \ref{equ:inversediffmean} and \ref{equ:inversediffvar}, and subtract it from the current state to obtain $\vx_{t-1}$. After arriving at $t=0$, we apply the clamping trick~\cite{li2022diffusion} to replace the values of $\vx_{0}$ with its closest word embeddings, and then decode the discrete tokens from $\vx_{0}$.

\paragraph{Training Phase} To train the diffusion model for sequence-to-sequence tasks, we first convert the target sequence $\vy = \{\evw^y_1, \evw^y_2, \dots, \evw^y_n\}$ into a continuous state $\vx_0$ using the embedding function with a additional Gaussian noise permutation, which can be expressed as:
\begin{equation}
\footnotesize
q(\vx_0 | \vy) = \mathcal{N}\left(\vx_{0} ; \text{Emb}(\vy),  \beta_{0} \mathbf{I}\right)
\label{equ:dis2continue}
\end{equation}
where $\text{Emb}(\cdot)$ is embedding function, $\beta_{0}$ represents the scaling of variance at time step $t=0$.
Then we apply the forward diffusion process~(\eqref{equ:forwarddiff}) to obtain the state $\vx_t$ at any step $t$ as a function of $\vx_0$, as shown in equation:
\begin{equation}
\footnotesize
q(\vx_t | \vx_0) = \mathcal{N}\left(\vx_{t} ; \sqrt{\bar{\alpha}_{t}}\vx_{0}, \sqrt{1-\bar{\alpha}_t}\mathbf{I}\right)
\label{equ:forwarddiffxt}
\end{equation}
where $\bar{\alpha}_{t} = \prod_{i=1}^t \alpha_i$.
In the training phase, we sample a random step $t$ to calculate $\vx_t$, and then use the denoising architecture to predict the noise for that step, based on the cross-attention with the source representation $\mH_s$. The mean and variance of the predicted noise are given by equations \ref{equ:xtpredictmean}: 
\begin{equation}
\footnotesize
\mu^{t-1}_{\theta}=\frac{1}{\sqrt{\alpha_{t}}}\left(\vx_{t}-\frac{\beta_{t}}{\sqrt{1-\bar{\alpha}_{t}}} z_{\theta}\left(\vx_{t}, t, \mH_s\right)\right)
\label{equ:xtpredictmean}
\end{equation}
where $z_{\theta}$ is the output of the denoising architecture and $\theta$ are its parameters.
The training objective is to minimize the squared error between the predicted and true noise, as well as the reconstruction error between $\vx_0$ and the target embeddings, as expressed in equation \ref{equ:seq2seqdiffobj}:
\begin{align} \label{equ:seq2seqdiffobj}
\footnotesize
\mathcal{L}_{\text {\emph{s2s}}}= 
\underset{q\left(\vx_{0:T} \mid \vy \right)}{\mathbb{E}}&[
\sum_{t=1}^{T} \left\|\mu^{t-1}_{\theta}-\hat{\mu}_{t-1}\right\|^{2} \\\nonumber
&+ \left\|\text{Emb}(\vy)-\mu^{0}_{\theta}\right\|^2 - \log p_{\theta}(\vy|\vx_0) ]
\end{align}
where $p_{\theta}(\vy|\vx_0) = \prod^{n}_{i=1}p_{\theta}(\evw_i^y | \vx_0)$, represents mapping the continuous latent variable $\vx_0$ into the discrete space token $\evw^y_i$.


\subsection{Pre-training \ours{}}
\label{subsec.pretraining}


Diffusion models have great potential for natural language generation (NLG) due to their ability to produce diverse outputs. However, they have been largely overlooked in NLG because of their slow convergence and low quality compared to autoregressive models. In this section, we address these challenges by pre-training a diffusion language model and introducing a novel pre-training task tailored for it. The novel pre-training task we propose is called \textit{continuous paragraph denoise} (CPD). CPD aims to train the model to predict the noise added to a continuous paragraph in the current diffusion step, given the paragraph and its surrounding context.


Specifically, given a document $\vd = \{\evw^d_1, \evw^d_2, \dots, \evw^d_l\}$ with $l$ words, we randomly select a paragraph $\vp = \{\evw^p_1, \evw^p_2, \dots, \evw^p_m\}$ from $\vd$, where $m = \lfloor\gamma * l\rfloor$ is the paragraph length and $\gamma$ is a predefined ratio. We mask the paragraph in the document with a special token ([MASK]), and feed the masked document $\vd' = \{\evw^{d'}_1, \evw^{d'}_2, \dots, \text{[MASK]}, \dots, \evw^{d'}_{l-m}\} $ to the \ours{} encoder. We also apply the forward diffusion process to the paragraph $\vp$ and obtain a noisy version $\vx_t$ at a random step $t$, and feed it to the \ours{} denoising architecture. The denoising architecture then uses the cross-attention with the source representation $\mH_s$ to predict the noise for the current step, using equations \ref{equ:xtpredictmean}. In summary, the pre-training objective of CPD is to minimize the same loss as in equation \ref{equ:seq2seqdiffobj}, except that $\vy$ is replaced by $\vp$ and $\vx_0$ is the embedded paragraph with noise.


Through this pre-training task, the diffusion model can enhance its semantic understanding of the continuous text and its denoising ability at each diffusion step. Moreover, the CPD task is self-supervised and does not rely on external labelled data sources, so it can fully exploit the information in the original pre-trained corpus. 

\section{Experiments and Results}\label{sec.experiment.structure}

\begin{table*}[h]
\centering
\caption{Results of Semi-NAR, NAR and AR on \textsc{XSum}. Index \textbf{OVERALL} represents the average value of \textbf{ROUGE-1}, \textbf{ROUGE-2} and \textbf{ROUGE-L}. It should be noted that \ours{} belongs to Semi-NAR.}
\small 
\begin{tabular}{l|c|cccc}
\hline
\multirow{2}{*}{\textbf{Methods}} &\multirow{2}{*}{\textbf{Pattern}} &  \multicolumn{4}{c}{\textbf{\textsc{XSum}}} \\                   
 &  & \textbf{ROUGE-1} & \textbf{ROUGE-2} & \textbf{ROUGE-L} & \textbf{OVERALL}   \\ \hline                                    

NAT~\cite{gu2017non} &\multirow{6}{*}{NAR} &  24.0 & 3.9 & 20.3 & 16.1  \\ 
 iNAT~\cite{lee2018deterministic} & & 24.0 & 4.0 & 20.4 & 16.1  \\ 
 CMLM~\cite{ghazvininejad2019mask} &  &  23.8 & 3.6 & 20.2 & 15.9   \\
 LevT~\cite{gu2019levenshtein}  & & 24.8 &  4.2 & 20.9 & 16.6   \\
BANG~\cite{qi2021bang} & & 32.6 & 9.0 & 27.4 & 23.0 \\
ELMER~\cite{DBLP:journals/corr/abs-2210-13304} & & \textbf{38.3} & \textbf{14.2} & \textbf{29.9} & \textbf{27.5} \\ \hline
LSTM~\cite{GreffSKSS17} &\multirow{6}{*}{AR} & 25.1 & 6.9 & 19.9 & 17.3 \\ 
Transformer~\cite{vaswani2017attention} & &  30.7    & 10.8 & 24.5 & 22.0  \\ 
MASS~\cite{SongTQLL19} &  & 39.7 &  17.2 & 31.9  & 29.6   \\
BART~\cite{lewis2019bart} & &  39.8 & 17.2 & 32.2  & 29.7  \\
ProphetNet~\cite{qi2020prophetnet} &  & 39.9 &  17.1 & 32.1  & 29.7  \\
BANG~\cite{qi2021bang} & & \textbf{41.1} & \textbf{18.4} & \textbf{33.2} & \textbf{30.9 } \\  \hline

InsT~\cite{stern2019insertion} & \multirow{7}{*}{Semi-NAR}&  17.7 & 5.2 & 16.1  & 13.0  \\ 
iNAT~\cite{lee2018deterministic}  & & 27.0 & 6.9 & 22.4 & 18.8   \\ 
CMLM~\cite{ghazvininejad2019mask} & & 29.1 & 7.7 & 23.0  & 20.0  \\
LevT~\cite{gu2019levenshtein}  & & 25.3 & 7.4 & 21.5 & 18.1   \\
BANG~\cite{qi2021bang}  & &34.7 & 11.7 & 29.2 & 25.2  \\
\ours{} (w/o pre-train) & & 38.9 & 17.5  & 31.0 
 & 29.1\\
\ours{} & & \textbf{42.9} & \textbf{21.4} & \textbf{35.1} & \textbf{33.2}\\ \hline

\end{tabular}
\label{table.result.XSum}
\end{table*}

\begin{table*}[t]
\small
\centering
\setlength\tabcolsep{5pt}
\caption{The main results on \textsc{CNN/DailyMail} and \textsc{Gigaword}.}
\begin{tabular}{l|ccc|ccc}
\hline
\multirow{2}{*}{\textbf{Method}} & \multicolumn{3}{c|}{\textbf{\textsc{CNN/DailyMail}}} & \multicolumn{3}{c}{\textbf{\textsc{Gigaword}}} \\
 & \textbf{ROUGE-1} & \textbf{ROUGE-2} & \textbf{ROUGE-L} & \textbf{ROUGE-1} & \textbf{ROUGE-2} & \textbf{ROUGE-L} \\ \hline

NAG-BERT~\cite{SuCWVBLC21} & - & - & - & 35.1 & 16.5 & 33.3 \\
LSTM~\cite{GreffSKSS17} & 37.3 & 15.7 & 34.4 & 34.2 & 16.0 & 31.8 \\
Transformer~\cite{VaswaniSPUJGKP17} & 39.5 & 16.7 & 36.7 & 37.1 & 18.4 & 34.5 \\
BART~\cite{lewis2019bart} & 41.3 & 19.4 & 38.1 & 38.6 & 19.5 & 35.7 \\
MASS~\cite{SongTQLL19} & 42.1 & 19.5 & 39.0 & 38.7 & 19.7 & 35.9 \\
ProphetNet~\cite{qi2020prophetnet} & 42.5 & 19.7 & 39.5 & 38.9 & 19.9 & 36.0 \\ \hline
\ours{} (w/o pre-train) & 43.8 & 20.6 & 41.2 & 43.7 & 23.3 & 40.8 \\
\ours{} & \textbf{45.6} & \textbf{23.2} & \textbf{43.1} & \textbf{45.7} & \textbf{25.8} & \textbf{42.9} \\ \hline
\end{tabular}
\label{table.result.cnndm}
\end{table*}
\begin{table*}[t]
\small
\centering
\setlength\tabcolsep{12pt}
\caption{The main results on \textsc{CommonGen}.}
\begin{tabular}{l|cccccc}
\hline
\multirow{2}{*}{\textbf{Method}} & \multicolumn{6}{c}{\textbf{\textsc{CommonGen}}} \\
 & \multicolumn{2}{c}{\textbf{ROUGE-2/L}} & \multicolumn{2}{c}{\textbf{BLEU-3/4}} & \textbf{CLDEr} & \textbf{SPICE} \\ \hline
bRNN-CopyNet~\cite{GuLLL16} & 7.7 & 27.8 & 12.6 & 7.1 & 5.1 & 13.4 \\
Trans-CopyNet~\cite{lin2019commongen} & 8.6 & 27.9 & 12.5 & 7.6 & 4.7 & 12.9 \\
MeanPooling-CopyNet~\cite{lin2019commongen} & 9.7 & 31.2 & 12.5 & 7.1 & 5.2 & 15.2 \\
LevT~\cite{GuWZ19} & 10.6 & 31.9 & 21.5 & 12.7 & 7.5 & 16.8 \\
ConstLeven~\cite{SusantoCT20} & 11.8 & 33.0 & 20.9 & 11.3 & 10.8 & 20.1 \\
T5-Base~\cite{RaffelSRLNMZLL20} & 14.6 & 34.6 & 28.8 & 18.5 & 9.4 & 19.9 \\ \hline
\ours{} (w/o pre-train) & 14.6 & 36.0 & 21.0 & 12.5 & 8.1 & 20.6 \\
\ours{} & \textbf{26.2} & \textbf{43.9} & \textbf{29.5} & \textbf{19.6} & \textbf{10.3} & \textbf{23.4} \\ \hline
\end{tabular}
\label{table.result.commongen}
\end{table*}

In this section, we will introduce the details of \ours{} pre-training, the data setting, and show extensive experimental results on various NLG downstream tasks.

\subsection{\ours{} Pre-training}

\paragraph{Model Framework} Our model uses a 6-layer transformer as the encoder, and a 6-layer cross attention transformer as denoising architecture. In particular, in denoising architecture, we use the randomly embedding function to map discrete token into continuous variable. We set latent variable dim to 768 and embedding dim to 128.

\paragraph{Pre-training Data}
Resent works has shown that pre-training on large scale corpus can improve the performance of the model on downstream tasks~\cite{lewis2019bart, qi2020prophetnet}, which is also applicable to \ours{} based on diffusion model. Following BART~\cite{lewis2019bart}, we use pre-training data consisting of 160Gb of news, books, stories, and web text. We segment sentences belonging to different chapters, and ensure that the input text length does not exceed 512.

\paragraph{Pre-training Setting}
We use the CPD task mentioned in \cref{subsec.pretraining} to pre-train \ours{} on large-scale corpus. The proportion of continuous paragraph $\gamma$ sets to 30\%, hence, for the 512 length input, the target length is 153. We randomly extract 153 length targets from the text input, and leave [MASK] token at the extracted position. In the training process, we use Adam optimizer~\cite{KingmaB14} with learning rate 1e-4, and we set the batch size to 512. We pre-trained our model on 8 × 40GB NVIDIA A100 GPUs with 5 million steps, lasting for 50 days. 
In the fine-tuning phase, we use the final pre-training model checkpoint to conduct fine-tuning on various downstream tasks.

\subsection{Fine-tune on Downstream Tasks}

In order to verify the effectiveness of pre-training on \ours{} based on diffusion model, we fine-tune and verify the effect of \ours{} on a variety of downstream tasks.
Through the above task, we can prove that the pre-trained \ours{} can quickly adapt to different types of NLG tasks without long time training like other diffusion models. 

\paragraph{Text Summarization}
As an important task in the NLG field, text summarization aims to summarize long documents into fluent short texts. In the experiment, we selected three widely used datasets: (a) \textsc{Gigaword} corpus~\cite{RushCW15}, (b) \textsc{CNN/DailyMail}~\cite{HermannKGEKSB15}, and (c) \textsc{XSum}~\cite{NarayanCL18}. In the process of fine-tuning, we set the learning rate to 5e-5 and the 120K training steps for all three dataset. In the inference process, we randomly sample 10 Gaussian noises for iteration denoising, and use the highest score as the final generated result. For different sample number, please refer to \cref{sec.appendix.samplenum}. During evaluation, we following the existing work~\cite{lewis2019bart, qi2020prophetnet}, reporting F1 scores of \textbf{ROUGE-1}, \textbf{ROUGE-2}, and \textbf{ROUGE-L} on test set.


\paragraph{Common Sense Generation}
Common sense generation tasks require the model have the ability of generative commonsense reasoning. Specifically, given a series of common sense concepts, the model needs to generate coherent statements based on these concepts that adhere to real-world scenarios. We select the widely used dataset \textsc{CommonGen}~\cite{lin2019commongen} to evaluate whether \ours{} has good creativity and reasoning ability in natural language generating. In the fine-tuning phase, we set the learning rate to 1e-4 and train 10k steps in total. Finally, we randomly sampled 10 gaussian noises and selected the best sample as the final result. Referring to the previous work~\cite{lin2019commongen}, we reported the indicators including F1 scores of \textbf{ROUGE-2/L}, \textbf{BLEU-3/4}, \textbf{CLDEr}, and \textbf{SPICE}.


\subsection{Baselines}

We compare \ours{} with the baselines of several mainstream methods. Specifically, these methods can be divided into two groups. The first group is the NAR model, including NAT~\cite{gu2017non}, iNAT~\cite{lee2018deterministic}, NAG-BERT~\cite{SuCWVBLC21},  CMLM~\cite{ghazvininejad2019mask}, LevT~\cite{gu2019levenshtein}, ConstLeven~\cite{SusantoCT20}, BANG~\cite{qi2021bang}, ELMER~\cite{DBLP:journals/corr/abs-2210-13304} and InsT~\cite{stern2019insertion}. Among them, InsT, iNAT, CMLM, LevT, ConstLeven and BANG can also be used in Semi-NAR, which can optimize the generation quality through multiple NAR iterations. It is worth noting that \ours{} also belongs to the Semi-NAR model.

The second group is AR model, the model of encoder-decoder structure including LSTM~\cite{GreffSKSS17}, Transformer~\cite{VaswaniSPUJGKP17}, bRNN-CopyNet~\cite{gu2016incorporating}, Trans-CopyNet~\cite{lin2019commongen}, MeanPooling-CopyNet~\cite{lin2019commongen} without pre-training, and strong baselines MASS~\cite{SongTQLL19}, BART~\cite{lewis2019bart}, T5~\cite{RaffelSRLNMZLL20}, BANG~\cite{qi2021bang}, and ProphetNet~\cite{qi2020prophetnet} with large scale pre-training. For large scale pre-training models mentioned above, we select the base version of the model, which is equivalent to the total number of \ours{} parameters. 






\subsection{Main Results}

We present the results of \ours{} and the baselines on \textsc{XSum}, \textsc{CNN/DailyMail}, \textsc{Gigaword}, \textsc{CommonGen} in \cref{table.result.XSum}, \cref{table.result.cnndm}, and \cref{table.result.commongen}. Our results demonstrate that the pre-trained \ours{} is a powerful NAR model for text generation. Especially on the \textsc{Xsum} dataset, \ours{} outperforms other NAR and Semi-NAR methods by a large margin, and on all three text summarization datasets, \ours{} achieves comparable quality to the pre-trained AR model. 
In addition, \ours{} shows creativity and logic in common sense generation tasks. On \textsc{CommonGen}, \ours{} surpasses other baseline models, including T5 which has been pre-trained on a large-scale corpus.


We also compare the pre-trained \ours{} and \ours{} trained from scratch (w/o pre-train). As shown in \cref{table.result.XSum} and \cref{table.result.cnndm}, pre-training significantly improves the \textbf{ROUGE-1}, \textbf{ROUGE-2}, \textbf{ROUGE-L} scores of \ours{} on the three text summarization datasets. Similarly, the results on \textsc{CommonGen} in \cref{table.result.commongen} indicate that pre-training enhances the performance of \ours{} on this task. These results confirm the effectiveness of our pre-training method.


\subsection{Generate Diversity Comparison}
\begin{table*}[]
\center
\setlength\tabcolsep{10pt}
\caption{\textbf{SELF-BLEU} score of BART and \ours{} generated results. For each data sample, we use BART and \ours{} to generate 10 summaries to evaluate diversity.}
\begin{tabular}{l|c|ccc}
\hline
\textbf{Model} & \textbf{Generate Method} & \textbf{\textsc{Xsum}} & \textbf{\textsc{CNN/DailyMail}} & \textbf{\textsc{Gigaword}} \\ \hline
\multirow{6}{*}{BART} & Greedy Search & 100.0 & 100.0 & 100.0 \\
& Beam Search & 93.4  & 96.2 & 90.2\\
& Diverse Beam Search & 75.6  & 84.1 & 71.8\\
& Typical Sample & 76.9  & 84.6 & 80.1\\
& Top-k Sample & 80.2  & 85.2 & 82.6\\
& Nucleus Sample & 79.1  & 83.5 & 79.4\\\hline
\ours{} & Diffusion & \textbf{29.3}  & \textbf{37.6} & \textbf{39.9}\\ \hline
\end{tabular}
\label{tab:diversityresult}
\end{table*}
\begin{table*}[]
\center
\setlength\tabcolsep{10pt}
\caption{Large language model evaluation on three summarization benchmarks.}
\begin{tabular}{c|cc|cc|cc}
\hline
 \textbf{Method} & \multicolumn{2}{c|}{\textbf{\textsc{XSum}}} & \multicolumn{2}{c|}{\textbf{\textsc{CNN/DailyMail}}} & \multicolumn{2}{c}{\textbf{\textsc{Gigaword}}} \\ 
 & BART & \ours{} & BART & \ours{} & BART & \ours{}\\ \hline
Average Summary Score & 2.69 & 2.58 & 2.96 & 2.90 & 2.58 & 2.46 \\
Average High-quality Summary & 6.91 & 5.95 & 9.66 & 9.04 & 5.99 & 4.99 \\ \hline
\end{tabular}
\label{tab:davenqiresult}
\end{table*}
With the emergence of the diffusion based model such as \ours{}, the advantages of text generation in diversity will be gradually valued. In this experiment, we will use both quantitative metrics and qualitative examples to show the richness of \ours{} in text generation.



To measure the diversity of \ours{} generation, we use \textbf{SELF-BLEU} as the metric. The lower the \textbf{SELF-BLEU} score, the more diverse the generated texts are. For comparison, we use BART, a state-of-the-art autoregressive model, which is pre-trained on large scale corpora. For BART, we apply different decoding methods of autoregressive models, such as greedy search, beam search~\cite{journals/corr/abs-2204-09269}, diverse beam search (diversity strength = 0.8)~\cite{journals/corr/VijayakumarCSSL16}, typical sampling ($\tau = 1.2$)~\cite{journals/corr/abs-2202-00666}, top-k sampling ($k = 50$)~\cite{LewisDF18}, and nucleus sampling ($p = 0.92$)~\cite{HoltzmanBDFC20}.
These decoding methods can generate multiple texts from the same source sequence. In this experiment, we generate 10 different target sequences for each source sequence using \ours{} and BART. Then we use the 10 summaries generated from \textsc{XSum}, \textsc{CNN/DailyMail}, and \textsc{Gigaword} to calculate the \textbf{SELF-BLEU} scores.
 

As shown in \cref{tab:diversityresult}, although the diversity of autoregressive generation can be slightly improved by using diverse beam search or some sampling methods with BART, the improvement is not significant. On the other hand, the diversity of generation is greatly enhanced by using the \ours{}. The large gaps in \textbf{SELF-BLEU} indicate that \ours{} can generate more diverse texts, not just varying a few words.


To complement the quantitative metrics, we also provide a case study in \cref{sec.appendix.casestudy} to analyze the quality of the texts generated by BART and \ours{}. We find that the autoregressive generation method can produce high-quality texts when there is only one output, but when generating multiple outputs, even with different decoding methods, it is hard to increase its diversity, and there may be many repeated prefixes. In contrast, the diffusion generation method can maintain the quality of generation while offering rich diversity.


However, it may not be fair to compare \ours{} directly with the single reference to prove that \ours{} can achieve diversity without compromising quality. Therefore, we design a new evaluation method. We use \texttt{text-davinci-003} version of InstructGPT~\cite{ouyang2022training}, which is based on the large language model (LLM) GPT-3.5, to score our generated texts, that is, to evaluate the quality of the generated summaries.
Specifically, we first obtain the sample set (10 summaries generated by BART using diverse beam search and 10 summaries generated by \ours{}), and design a prompt to input into \texttt{text-davinci-003} to score the generated summaries, while counting the number of high-quality summaries within the 10 summaries generated by BART and \ours{} respectively.
We conduct the experiment on the three different text summarization datasets and use two evaluation methods, \textit{Average Summary Score} represents the average score given by \texttt{text-davinci-003}, ranging from 1 to 3, and \textit{Average High-quality Summary} represents the average number of high-quality summaries in 10 samples, ranging from 0 to 10. For more detailed experimental settings, please refer to \cref{sec.appendix.LLMEval}.


As shown in \cref{tab:davenqiresult}, although \ours{}'s scores are slightly lower than BART's, according to the results in \cref{tab:diversityresult}, the diversity of samples generated by BART is much lower than \ours{}. Given the trade-off between diversity and quality, the score difference is within the acceptable range. Moreover, the result of \textit{Average High-quality Summary} shows that there are still enough high-quality summaries in the case of high diversity. Such advantages of \ours{} deserve our attention and further exploration in our future work.


\begin{table}[t]
\small
\centering
\caption{Effect of pre-training step, on \textsc{XSum}. The result is the optimal value of 5 Gaussian samples.}
\setlength\tabcolsep{3pt}
\begin{tabular}{l|ccc}
\hline
\textbf{Model}  & \multicolumn{1}{l}{\textbf{ROUGE-1}} & \multicolumn{1}{l}{\textbf{ROUGE-2}} & \multicolumn{1}{l}{\textbf{ROUGE-L}} \\ \hline
w/o pre-train & 37.3 & 15.3 & 29.4 \\
\ours{}(100w) & 39.4 & 17.1 & 31.5 \\
\ours{}(200w) & 40.4 & 18.2 & 32.5 \\
\ours{}(300w) & 40.6 & 18.5 & 32.8 \\
\ours{}(400w) & 40.9 & 18.7 & 33.0 \\
\ours{}(500w) & 41.2 & 19.1 & 33.4 \\ \hline
\end{tabular}
\label{tab:pretrainstep}
\end{table}
\begin{table}[t]
\small
\centering
\caption{Effect of the proportion of continuous paragraphs, on \textsc{XSum}. The result is the optimal value of 5 Gaussian samples.}
\setlength\tabcolsep{5pt}
\begin{tabular}{c|ccc}
\hline
\textbf{CPD Proportion}  & \multicolumn{1}{l}{\textbf{ROUGE-1}} & \multicolumn{1}{l}{\textbf{ROUGE-2}} & \multicolumn{1}{l}{\textbf{ROUGE-L}} \\ \hline
$\gamma = 15\%$ & $40.24_{\pm0.03}$ & $18.03_{\pm0.01}$ & $32.30_{\pm0.02}$ \\
$\gamma = 20\%$ & $40.14_{\pm0.10}$ & $17.90_{\pm0.04}$ & $32.21_{\pm0.08}$ \\
$\gamma = 25\%$ & $40.24_{\pm0.10}$ & $18.12_{\pm0.04}$ & $32.37_{\pm0.04}$ \\
$\gamma = 30\%$ & \bm{$40.37_{\pm0.04}$} & \bm{$18.20_{\pm0.02}$} & \bm{$32.58_{\pm0.05}$} \\
$\gamma = 35\%$ & $40.29_{\pm0.06}$ & $18.18_{\pm0.02}$ & $32.45_{\pm0.06}$ \\
$\gamma = 40\%$ & $40.15_{\pm0.08}$ & $17.87_{\pm0.07}$ & $32.12_{\pm0.09}$ \\ \hline
\end{tabular}
\label{tab:maskpro}
\end{table}
\begin{table}[t]
\small
\centering
\caption{Difference between uniform time schedule sample(UI) and loss aware sample(LA), on \textsc{XSum}. The result is the optimal value of 5 Gaussian samples.}
\setlength\tabcolsep{8pt}
\begin{tabular}{l|ccc}
\hline
\textbf{Method}  & \multicolumn{1}{l}{\textbf{ROUGE-1}} & \multicolumn{1}{l}{\textbf{ROUGE-2}} & \multicolumn{1}{l}{\textbf{ROUGE-L}} \\ \hline
$\gamma = 15\%, \text{UI}$ & $40.24_{\pm0.03}$ & $18.03_{\pm0.01}$ & $32.30_{\pm0.02}$ \\
$\gamma = 15\%, \text{LA}$ & $40.06_{\pm0.04}$ & $17.90_{\pm0.02}$ & $32.17_{\pm0.05}$ \\
$\gamma = 30\%, \text{UI}$ & $40.37_{\pm0.04}$ & $18.20_{\pm0.02}$ & $32.58_{\pm0.05}$ \\
$\gamma = 30\%, \text{LA}$ & $40.18_{\pm0.03}$ & $17.94_{\pm0.02}$ & $32.24_{\pm0.02}$ \\ \hline
\end{tabular}
\label{tab:tsample}
\end{table}

\subsection{Impact of Pre-training Steps}

Our pre-training method and the diffusion model itself are both designed to achieve long-term convergence and unlimited potential, but they also require a large amount of pre-training time. Here we investigate how the pre-training steps affect the performance of our model compared with a non-pre-trained \ours{} on the \textsc{XSum} dataset. We fine-tune the checkpoints obtained at 1 million step intervals from pre-training and evaluate them using 5 random Gaussian noises, selecting the highest score as the final result. As shown in \cref{tab:pretrainstep}, pre-training for only 1 million steps can significantly improve the quality of generation over the non-pre-trained \ours{}. Moreover, we can see from the results that pre-training continues to steadily boost the performance of the \ours{} on the downstream task as the pre-training steps increase.

\subsection{Impact of Pre-training Parameters}

In this subsection, we examine the effect of important pre-training parameters on the pre-training performance. First, in the unsupervised pre-training method CPD, we need to explore how the proportion of continuous paragraphs $\gamma$ influences the pre-training performance. We vary the value of $\gamma$ from 15\% to 40\% (with 5\% intervals) and conduct 2 million steps of pre-training for each value. After the pre-training, we evaluate the pre-training effect by fine-tuning on \textsc{Xsum}. For a rigorous evaluation, we sample 5 Gaussian noises, repeat the experiment 5 times with different random seeds, and report the mean and standard deviation of the results, each time choosing the highest score as the final result. As shown in \cref{tab:maskpro}, too large or too small values of $\gamma$ lead to instability and poor performance of the pre-trained model. Pre-training is more stable and effective when $\gamma=30\%$.


Second, we investigate the time step sampling method used in the pre-training. Before each training step, we need to sample a time step as part of the model input. The existing two common time step sampling methods are uniform sample and loss aware sample. The former assigns equal probabilities to each time step, while the latter updates the sampling weights according to the training loss, so that more important time steps have higher chances of being sampled. In the experiment, we use these two sampling methods, test them on two different values of $\gamma$ (15\% and 30\%), and perform a rigorous evaluation similar to the previous experiment. As shown in \cref{tab:tsample}, we observe that under 2 million steps of pre-training, the uniform sample outperforms the loss aware sample for different values of $\gamma$. Intuitively, although the loss aware sample can speed up the convergence of the diffusion model, we hope that the model can learn sufficient knowledge at each time step during the pre-training, so that it can converge faster and perform better on downstream tasks.

\subsection{Impact of Diffusion Time Step}
\begin{figure}[t]
\centering
\includegraphics[width=0.4\textwidth]{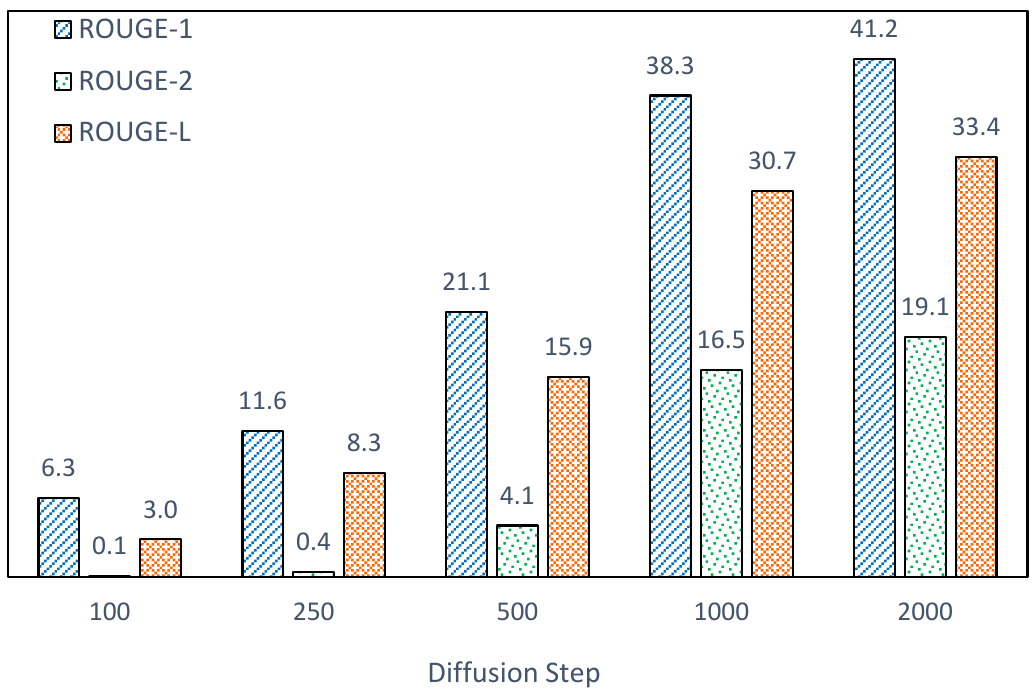}
\caption{Effect of different diffusion steps on text generation quality, on \textsc{XSum}. The result is the optimal value of 5 Gaussian samples.}
\label{fig:timestep}
\end{figure}


The number of diffusion time steps has a great impact on the quality of generation. We explore how the \ours{} performs under different numbers of inverse diffusion steps on the \textsc{XSum} dataset. Assuming that the total number of diffusion steps $T=2000$, we set the interval step of inverse diffusion to 1, 2, 4, 8, 20, and the corresponding numbers of inverse diffusion steps are 2000, 1000, 500, 250, 100. In this experiment, we sample 5 Gaussian noises and choose the best denoising result. As shown in \cref{fig:timestep}, we can clearly see that when the number of inverse diffusion steps is small, the quality of generation with \ours{} deteriorates significantly. As the number of inverse diffusion steps increases to 1000, the generation quality of \ours{} becomes stable.

\section{Related Work}

\subsection{Large Scale Pre-training Language Models}
Recently, a major breakthrough has been made in the model of pre-training on large scale corpus. 
As unidirectional language models, GPT~\cite{radford2018improving}, GPT2~\cite{radford2019language} modeling the text based on left-to-right, and predict the next token according to the token appearing on the left. 
At the same time, bidirectional language models, which uses bidirectional encoder to model text, can obtain better context sensitive representation, such as BERT~\cite{DevlinCLT19} and RoBERT~\cite{journals/corr/abs-1907-11692}. RoBERT optimizes pre-training tasks compared to BERT, both of which significantly improve the ability of natural language understanding. 
In order to improve the performance of the large scale pre-training model in natural language generation, some works has designed pre-training tasks based on the standard framework of sequence-to-sequence. 
MASS~\cite{SongTQLL19} lets the model predict the short masked token span step by step, while ProphetNet~\cite{qi2020prophetnet} predict more words in each step to ease local over fitting. 


\subsection{Diffusion Models for Text}

In recent years, diffusion model has achieved great success in the domains of image generation~\cite{journals/corr/abs-2204-06125, journals/corr/abs-2205-11487, RombachBLEO22}. Because of its amazing generation quality, some works apply diffusion model in text generation domains. Diffsuion-LM~\cite{li2022diffusion} maps discrete tokens into continuous latent variable, achieving more complex controllable text generation through continuous diffusion. In the field of text revision where non-autoregressive method is widely used, DiffusER~\cite{reid2022diffuser} also uses the diffusion model to implement the edit based generative processes. DiffuSeq~\cite{gong2022diffuseq} achieves conditional text generation with a new method which controlled information is also involved in the diffusion process.
Different from the above work, we build a novel language model based on the diffusion model for the first time, using the standard enocoder-decoder framework. For our best knowledge, we are the first to adopt large scale pre-training on the language model based on the diffusion model.

\section{Conclusion}

In this paper, we have presented a novel diffusion language model \ours{}, which leverages a large scale corpus for pre-training. Our model adopts a sequence-to-sequence framework, where a bidirectional encoder encodes the source sequence and a denoising decoder predicts and removes noise from the target sequence in a non auto-regressive fashion. This design allows us to generate diverse text by gradually refining the output from a noisy initial state. Moreover, we have introduced a new pre-training method called \textit{continuous paragraph denoise}, which aims to denoise whole paragraphs as the target sequence. Our experiments on various NLG tasks demonstrate that \ours{} can produce high-quality and diverse text, and validate the benefits of pre-training our diffusion model on a large scale corpus.

\newpage

\bibliography{main}
\bibliographystyle{icml2021}


\appendix
\section{Case Study}
\label{sec.appendix.casestudy}

\begin{table*}[]
\center
\small
\caption{Summary examples of BART and \ours{} generated results.}
\begin{tabular}{p{4.1cm}|p{12cm}}\toprule

\hline
%
%

\textbf{source sequence \uppercase\expandafter{\romannumeral1} \newline (abbreviated)} & Those who participated in the Aberdeen Children of the 1950s project, which saw all primary pupils aged seven to 12 surveyed by the Medical Research Council in 1962, have been contacted. They have been asked to take part in the Scottish Family Health Study. It aims to investigate why diseases such as cancer can run in families.Those recruited will have their health tracked, with the intention of creating a Scottish "bio-bank" containing genetic, medical and family history and lifestyle information.  The data gathered would help future research into the prevention, treatment and diagnosis of illnesses. \\ \hline
\textbf{\ours{} summaries \uppercase\expandafter{\romannumeral1} \newline (diffusion)} & 1. health information have been recruited by university school primary pupils to help improve their lives. \newline  2. a health project is to be recruited by learning for university researchers in scotland. \newline 3. scientists in aberdeen are to meet experts in scotland to get more health data for their children. \newline \\ \hline
\textbf{BART summaries \uppercase\expandafter{\romannumeral1} \newline (diversity beam search)} & 1. thousands of children from aberdeen are being recruited to help scientists investigate why diseases run in families. \newline 2. thousands of children in scotland have been asked to take part in a new project to study their health. \newline 3. thousands of children in scotland have been asked to take part in a new project to study their health..\\ \hline

\hline


\textbf{source sequence \uppercase\expandafter{\romannumeral2} \newline (abbreviated)} & Elin Jones is expected to lay out plans where some areas of Welsh forest could be transferred to the private sector or to not for profit organisations.But she has already ruled out the widespread sale of Welsh woodlands.Forestry Commission Wales said it would explore the feasibility of transfer to the private sector case by case.The minister told BBC Radio Wales she plans to "compensate" the public by buying new land for new planting or management if any forest was sold off on a case-by-case basis."I don't want any stagnancy in the forest estate. I want it to work for public benefit whether that's economic or environmental or access benefit," she said."It's my view there should be no reduction in the publicly owned estate and I have asked the Forestry Commission to look at how it can make that estate work harder, provide a better return for the public."Whether that's in terms of public access, in terms of environmental benefit in the production of renewable energy or biomass potential or also in terms of the economic return from that forestry estate." \\ \hline
\textbf{\ours{} summaries \uppercase\expandafter{\romannumeral2} \newline (diffusion)} & 1. the forestry minister is preparing to fill out its plan for some members of the public on wales' forests to be reduced. \newline 2. the forestry minister is picking forward plans to tackle some of the companies in wales to develop a boost in the management of forests. \newline 3. the forestry minister hopes to face plans continue on the future of wales' forest estate to be held to a growing and better access to the private sector.\newline \\ \hline
\textbf{BART summaries \uppercase\expandafter{\romannumeral2} \newline (diversity beam search)} & 1. the environment minister has said there should be no "stagnancy" in the size of the welsh forest estate. \newline 2. the forestry minister has said there should be no "stagnancy" in the size of the welsh forest estate. \newline 3. the future of wales' forests could be decided by the environment minister. \\ \hline
\hline


%


\end{tabular}
\label{tab:casestudyresult}
\end{table*}
\begin{table*}[]
\center
\caption{Example of prompt used by \texttt{text-davinci-003}.}
\begin{tabular}{p{15cm}}\toprule
\hline
\textbf{Prompt Input \bm{$>$}} Deborah Steel, who was 37 and ran the Royal Standard in Ely, was last seen in the early hours of 28 December. Her body has not been found. No further action will be taken against a 50-year-old and a 70-year-old, both from Ely, Cambridgeshire Police said .A 72-year-old man from Ely has been re-bailed until 17 February. Ms Steel's disappearance was recently reclassified from a long-term missing person inquiry to a murder inquiry by officers. \newline \newline
If the following sentences as summary of the above article, please assign an overall score. Scores range from 1 to 3, 1 represents bad, 2 represents neutral, 3 represents good. The output format is 'Score: 1'. \newline \newline
Two of the three men arrested by detectives investigating the disappearance of a Cambridgeshire pub landlady in 1997 have been released. \newline
\newline
\textbf{Output \bm{$>$}} Score: 3
\\ \hline
\end{tabular}
\label{tab:prompt}
\end{table*}

In the section\cref{sec.experiment.structure}, we have made a rigorous analysis of the quality and diversity of \ours{}. We hope that by comparing diffusion model with traditional autoregressive generation models, we can find the potential of diffusion model in natural language generation tasks. Nowadays, most of the excellent language generation models belong to autoregressive generation models, but at the same time, we also need some new ideas and generation paradigm to make natural language generation not limited to autoregressive. The ways of natural language generation need to be diversified to broaden researchers' thinking, just as the application of diffusion model in natural language generation can bring rich diversity to the generated text. We are excited that the diversity of the content generated by the diffusion model does not come from a large number of wrong words or unrelated texts, but from different sentence patterns and different information obtained from the original text. This shows us the future prospects of the diffusion model in the natural language generation task.

In order to more intuitively show the quality of the text generated by the diffusion model and the autoregressive generation model, we selected two samples from the text summarization dataset \textsc{XSum} in \cref{tab:casestudyresult}. For each sample, we used \ours{} and BART to generate three summaries respectively, of which the BART generation method is diversity beam search. For display purposes, the source sequence has been intercepted and abbreviated to reduce the length. 
It can be seen from the generated summaries that if we only observe one sentence of the generated summary, the summary generated by the autoregressive model BART is of good quality and is related to the content of the source sequence, the generated text is relatively fluent due to the autoregressive generation mode. 
But if we look at the three generated summaries, the text generated by the autoregressive model BART obviously has a lot of duplication. In the second example, "there should be no 'stagnancy' in the size of the Welsh forest estate" has repeated descriptions in two different summaries. In the first example, there are even two summaries that are almost identical. The difference is only one meaningless full stop at the end. 
Although we use the diversity beam search generation method when generating, it is difficult to let the autoregressive model jump out of the essence of iterative generation to generate creative text. 

In contrast, we can see that the summary generated by \ours{} may not be as fluent as BART due to the non-autoregressive generation mode if only from the quality of single sentence generation. Once multiple summaries are generated, we can observe \ours{}'s creative generating ability. In the first example, \ours{} describes the medical project in the source sequence from three related direction. Health information collection, project information and project objectives are mentioned in the generated summary. In the second example, "the forestry minister" mentioned a variety of measures on the forest industry, and the three summaries generated by \ours{} described different parts respectively, including the formulation of public plans, cooperation with Welsh companies to promote forest management, and the involvement of private enterprises in the forest industry. Compared with the single information "there should be no 'stagnancy' in the size of the Welsh forest estate" provided by BART, although BART also provides a concise summary, it is difficult to conclude that BART's summary is better, because people are accustomed to understanding some problems from multiple perspectives in practice, rather than a conclusive conclusion.

After analyzing the above examples, we can observe the great potential of new language model \ours{}. In the practical application of text generation, diversified generation results can be used in many scenarios. The unique generation method of diffusion model brings new ideas to text generation, and also lets us consider whether the single-label text generation really meets our needs. We do not need the diffusion model to be superior to the autoregressive generation model in all aspects. What we need is a new idea to bring more possibilities to text generation. We believe that the diffusion model can be widely used in text generation in the future.

\section{Large Language Model Evaluation}
\label{sec.appendix.LLMEval}
Recently, the large language model has been widely used in various tasks with its amazing performance. In this paper, we use the large language model to evaluate the quality of the generated summary. We select \texttt{text-davinci-003} as the evaluation model in our experiment, the most important thing is the construction of the prompt which will input into the model. 

As prompt example shown in \cref{tab:prompt}, we divide the prompt into three parts. The first part is the text of the original article, the middle part is the evaluation requirements, and the end part is the summary that needs to be evaluated. Finally, we can get output score through the large language model. For each summary, we need to organize the above prompt for the model input, but the evaluation requirements for each summary are the same. During the test, we asked the model to give the following score to the summary: 1 represents bad, 2 represents neutral, 3 represents goods. After getting the score of each summary, we will further count the number of high-quality samples in the 10 samples generated. Here, we define high-quality summary as summary with a score equal to 3. Finally, we can summarize and integrate all the scores obtained, and count the average summary score and the average number of high-quality summaries.

\section{Impact of Sample Number}
\label{sec.appendix.samplenum}
\begin{figure}[t]
\centering
\includegraphics[width=0.45\textwidth,height=2.0in]
{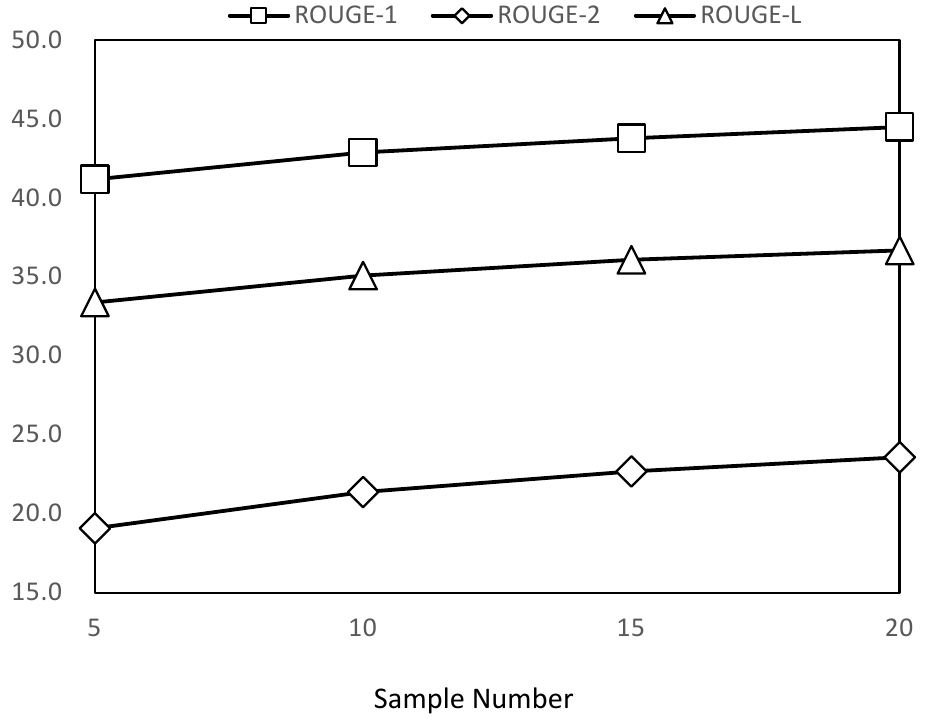}
\caption{Effect of sample number, on \textsc{XSum}.}
\label{fig:samplenum}
\end{figure}

Compared with the autoregressive model, \ours{} based on the diffusion model can generate more diverse texts in the inference phase, even under the guidance of the source sequence. Different Gaussian noises sampled during denoising can often lead to completely different generation results. This method is more flexible, but it is not conducive to the evaluation against a single reference answer. However, as the number of Gaussian noises sampled increases, the generated text has a higher probability of approaching the single reference answer, and the corresponding evaluation score is higher. To this end, we test the performance of the model on the test set under different numbers of samples on the \textsc{XSum} dataset. As shown in \cref{fig:samplenum}, we evaluate the results of 5, 10, 15 and 20 samples. We can observe that as the number of samples increases, the more likely the generated sample is to be similar to the original label. The improvement of the similarity is more noticeable in the early stage of increasing the number of samples.

\end{document}